%% file: main.tex
\documentclass[manuscript,screen,review=false]{acmart}
\usepackage{multirow}
\usepackage{amsfonts}
\usepackage{algorithm}
\usepackage[noend]{algpseudocode}
\usepackage{graphicx}
\usepackage{subcaption}
\AtBeginDocument{%
  \providecommand\BibTeX{{%
    \normalfont B\kern-0.5em{\scshape i\kern-0.25em b}\kern-0.8em\TeX}}}

\copyrightyear{}
\acmYear{}
\acmDOI{}

\acmConference[]{}
\acmBooktitle{}
\acmPrice{}
\acmISBN{}

\renewcommand\footnotetextcopyrightpermission[1]{} 

\begin{document}

\title{Predicting Terrorist Attacks in the United States using Localized News Data}

\author{Steven J. Krieg}
\affiliation{%
\institution{Lucy Family Institute for Data and Society, University of Notre Dame, skrieg@nd.edu}
}
\email{skrieg@nd.edu}

\author{Christian W. Smith}
\affiliation{%
\institution{Physical Sciences Inc., cwsmith@psicorp.com}
}
\email{cwsmith@psicorp.com}

\author{Rusha Chatterjee}
\affiliation{%
\institution{Physical Sciences Inc., rchatterjee@psicorp.com}
}
\email{rchatterjee@psicorp.com}

\author{Nitesh V. Chawla}
\affiliation{%
\institution{Lucy Family Institute for Data and Society, University of Notre Dame, nchawla@nd.edu*\thanks{* Corresponding author}}
}
\email{nchawla@nd.edu}

\renewcommand{\shortauthors}{Krieg et al.}

\begin{abstract}
Terrorism is a major problem worldwide, causing thousands of fatalities and billions of dollars in damage every year. Toward the end of better understanding and mitigating attacks that take place in the United States, we present a set of machine learning models that learn from localized news data in order to predict whether a terrorist attack will occur on a given calendar date and in a given state. The best model---a Random Forest that learns from a novel variable-length moving average representation of the feature space---achieves area under the receiver operating characteristic scores $> .667$ on four of the five states that were impacted most by terrorism between 2015 and 2018. Our key findings include that modeling terrorism as a set of independent events, rather than as a continuous process, is a fruitful approach---especially when the events are sparse and dissimilar. Additionally, our results highlight the need for localized models that account for differences between locations. From a machine learning perspective, we found that the Random Forest model outperformed several deep models on our multimodal, noisy, and imbalanced data set, thus demonstrating the efficacy of our feature representation method in such a context. Finally, we analyze factors that limit model performance, which include a noisy feature space and small amount of available data. These contributions provide an important foundation for the use of machine learning in efforts against terrorism in the United States and beyond.
\end{abstract}



\maketitle

\input{intro.tex}
\input{model.tex}
\input{results.tex}
\input{conclusion}

\section*{Acknowledgements}
This material is based upon work supported by the Army Contracting Command - Aberdeen Proving Ground, Edgewood Division under Contract No. W911SR-19-C-0007. Any opinions, findings and conclusions or recommendations expressed in this material are those of the author(s) and do not necessarily reflect the views of the Army Contracting Command - Aberdeen Proving Ground, Edgewood Division. The authors declare no competing interests.

\section*{Author Contributions}
C.W.S., R.C., and N.V.C. conceived of the research study; S.J.K. designed the predictive models; S.J.K., C.W.S, and R.C. evaluated the models and interpreted the results; S.J.K. drafted the manuscript; S.J.K., C.W.S., R.C., and N.V.C. reviewed the manuscript.

\section*{Data and Code Availability}
All data used in this study is publicly available from the corresponding references. All code will be published upon manuscript acceptance and is currently available upon request to the corresponding author.

\bibliographystyle{ACM-Reference-Format}
\bibliography{main}


\end{document}

%% file: intro.tex
\section{Introduction} \label{sec:intro}
Terrorism poses a significant threat to human livelihood around the world. 
According to the Institute for Economics and Peace (IEP), in 2018 alone incidents of terrorism killed 15,952 people and produced a \$33 billion (USD) global economic impact \cite{economics2019global}. Much of this activity occurs in the Middle East and other parts of Asia, leading many researchers to use machine learning to predict the evolution and spread of terrorism in such regions. However, other areas of the world---including the United States---are not exempt from the effects of terrorism. Between 2015 and 2018, the Global Terrorism Database (GTD) recorded 229 incidents of terrorism in the United States \cite{lafree2007introducing}. In the 2019 Global Terrorism Index, the IEP also noted that the United States has experienced one of the largest decreases in Positive Peace Index, a change that indicates increasing social disorder and greater risk of politically-motivated violence \cite{economics2019global}. Predicting such violence before it occurs could enable prevention strategies and save lives.

In this work, we present a machine learning model that uses localized news data to predict the occurrence of terrorist attacks in the United States. Specifically, our model predicts if an attack will occur on a given calendar date and in a given state. The model is trained on multimodal inputs, utilizing localized news data from the Global Database of Events, Language, and Tone (GDELT) to generate features for each location and date, and the Global Terrorism Database (GTD) to label each location and date pair. We consider data from 2015-2018 and focus our experiments on the five states in which the most attacks have been perpetrated: New York, California, Texas, Florida, and Washington. Our results show that a Random Forest model achieved area under the receiver operating curve (AUROC) scores $\geq 0.667$ on four of the five states, outperforming several other models---including neural networks. The Random Forest benefits significantly from a novel variable-length moving average method, which computes the statistically optimal window for each feature independently and thus helps prevent overfitting in our highly imbalanced, noisy, and small data setting. Our other key findings include that models should be localized (i.e., state models should be independently trained and evaluated) and that the characteristics of individual attacks (e.g., responsible group or weapon type) were not correlated with prediction success.

Several prior works address the problem of detecting terrorist activity, such as classifying pro-terrorism tweets \cite{abrar2019framework}. In this study, however, we are interested in the task of prediction rather than detection. Additionally, our task is to predict terrorist attacks or events, which is distinct from works that have sought to infer characteristics of an attack, such as the responsible group, after it has taken place \cite{tolan2015experimental, talreja2017terrorism, agarwal2019comparison, chaurasia2019global}. Few works have attempted to predictively model terrorist attacks, with the recent work of Python et al. being the noteworthy exception \cite{python2021predicting}. In this study, the authors train several machine learning models using prior terrorism data (from the GTD) in conjunction with other geographic and socioeconomic features to predict attacks at discrete spatiotemporal intervals. Despite promising results and thorough analysis at an impressive and global scale, we found two fundamental problems with their approach:
\begin{enumerate}
    \item Coarsely evaluating a model that represents a large region (e.g., West Africa) but is comprised of many smaller geographic cells overlooks the imbalanced spatial distribution of attacks. In this case, which is more complex than evaluation on a typical class-imbalanced dataset, measures like AUROC and area under the precision-recall curve (AUPRC) can be misleading. For example, using the data described in Table \ref{tab:attacksbystate}, we can create a terrorism model for the United States that predicts an attack will occur every day in five states (CA, NY, TX, FL, and WA) but that no attacks will occur in other locations. When coarsely evaluated across the entire United States, this model produces an AUROC of 0.733 and AUPRC of 0.468, both of which significantly outperform their respective baselines of 0.500 and 0.003---however, the predictions provide no value beyond reflecting that some regions have experienced a greater number of attacks. We avoid this problem by evaluating predictive performance for each region (state) individually.
    \item In order to make meaningful predictions at a granular time scale (e.g., will a terrorist attack take place during a given week), models must have inputs of a similar temporal granularity in order to differentiate between points in time. Many of the inputs utilized by Python et al. are relatively static, such as population density and gross product. The only temporally granular inputs were autoregressions based on local terrorist activity (i.e., whether there were recent terrorist attacks in the location of concern). While these autoregressions are valuable in their ability to model continuous terrorist activity, without other temporally relevant inputs the predictive models will be fundamentally limited in their ability to predict attacks at new or unusual locations. This shortcoming could also exacerbate the coarse evaluation problem described above.
\end{enumerate}

In contrast with the treatment of terrorism as a discrete prediction problem, a number of prior studies have modeled terrorism as a spatiotemporal process within a complex dynamical system. This body of work has generally been more concerned with modeling the evolution of terrorist organizations, understanding the escalation of violence, or quantifying the lethality or number of attacks within a geographical region and/or over a period of time \cite{johnson2011pattern, johnson2013particles, johnson2015modeling, ding2017understanding, vanderzee2018predicting, bang2018predicting, yang2019quantifying}. In another study, Python et al. focused on modeling the continuous phenomena (e.g., lethality and frequency) exhibited by terrorism in the Middle East \cite{python2019bayesian}. Li et al. similarly considered terrorism as ``an unstable system of interdependent events'' and treated it as a type of crime \cite{li2019longrange}. Other crime prediction models have had some success when applied to urban, high-crime areas like Chicago and New York City \cite{zhao2017modeling, kang2017prediction}, but these solutions do not necessarily translate to predicting terrorism, which is much more narrowly defined \cite{lafree2007introducing}. Although Li et al. apply their crime prediction model to terrorism the Middle East, neither theirs nor the other aforementioned works have considered terrorism in the United States, where attacks are much more sparse and the dynamical processes consequently more difficult to observe. According to the GTD, in 2017 Iraq experienced 2,466 terrorist attacks, 1,154 of which were perpetrated by a single organization: the Islamic State of Iraq and the Levant (ISIL). However, in the same year, the United States experienced just 65 attacks, only five of which were perpetrated by a known organization \cite{lafree2007introducing}. Models that rely on systematic patterns exhibited by organizations or interdependence between events are unlikely to succeed in this context.

Other related work includes the use of news data to predict social unrest. Qiao et al. use mentions of protests encoded in GDELT to predict social unrest events in Southeast Asia \cite{qiao2017predicting}. Galla et al. extract a set of longitudinal features from GDELT, which they use to predict social unrest events for a given location on a given date \cite{galla2018predicting}. We utilize a similar approach in extracting features from GDELT, which we describe in Section \ref{sec:methods}.

%% file: model.tex
\section{Materials and Methods} \label{sec:methods}

\begin{figure}[t]
    \centering
    \includegraphics[width=0.7\textwidth]{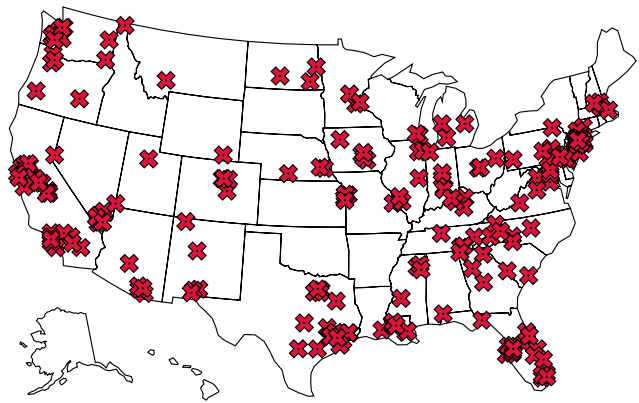}
    \caption{The locations of the 229 terrorist attacks perpetrated in the United States between Feb. 18, 2015, and Dec. 31, 2018, as recorded in the Global Terrorism Database.}
    \label{fig:gtdhistory}
\end{figure}

\begin{table}[]
    \centering
    \begin{tabular}{lrr}
         \textbf{State(s)} & \textbf{\# Attacks} & \textbf{Imbalance} \\ \hline
         CA, NY & 24 & .0170 \\ \hline
         TX & 18 & .0127 \\ \hline
         FL & 17 & .0120 \\ \hline
         WA & 14 & .0099 \\ \hline
         LA, MO & 7 & .0050 \\ \hline
         NV, PA & 6 & .0042 \\ \hline
         IN, NC, TN, VA & 5 & .0035 \\ \hline
         CO, IA, MS, NM & 4 & .0028 \\ \hline
         GA, IL, KY, MA, MN, ND, OH, OR & 3 & .0021 \\ \hline
         AZ, DC, MD, MI, NE, NJ, SC, UT, WI & 2 & .0014 \\ \hline
         CT, DE, ID, KS, MT, WY & 1 & .0007 \\ \hline
         AK, AL, AR, HI, ME, NH, OK, RI, SD, VT, WV & 0 & --- \\ \hline
    \end{tabular}
    \caption{Terrorist attack counts by state between Feb. 18, 2015, and Dec. 31, 2018, as recorded in the Global Terrorism Database. Imbalance is the proportion of the 1,413 days on which attacks occurred.}
    \label{tab:attacksbystate}
\end{table}

In this section we first formulate the problem of predicting the occurrence of a terrorist attack at a given location and time. We then describe our methods for data collection and preprocessing. Finally, we introduce the relevant learning algorithms and discuss our approaches to feature representation.

\subsection{Problem Formulation and Notation} \label{sec:problem} 
We formulate the task as a classification problem such that our goal is to learn a function $f : x \xrightarrow{} y$, where $x$ is a set of input features and the binary output $y$ represents the occurrence or non-occurrence of a terrorist attack at a particular location. We consider as locations the set of 51 states in the United States (including Washington D.C., excluding Puerto Rico and other non-state territories), each with $m = 862$ features observed over $n = 1,413$ calendar days from February 18, 2015 through December 31, 2018, inclusive (as limited by data availability). Let $X \in \mathbb{R}^{n \times m}$ be a feature matrix and $\vec{y} = \{0, 1\}^n$ be a label vector, such that $\vec{x}_i$ and $y_i$ denote the feature vector and label, respectively, that describe the $i^{th}$ day at the location of concern. Further, we use $X_{i,j}$ to represent the $i^{th}$ observation of the $j^{th}$ feature, $X_{i:k}$ to represent the $i^{th}$ through $k^{th}$ (inclusive) observations of all $m$ features, and $X_{*,j}$ to represent the column vector that contains all observations of the $j^{th}$ feature. Unless stated otherwise, we treat locations as independent such that each has a distinct feature matrix $X$ and label vector $\vec{y}$. Finally, we use square brackets $\left[ \right]$ to denote vectors.

\subsection{Data Collection} \label{sec:datacoll}
\subsubsection{Feature Extraction} \label{sec:featureextraction}

\begin{figure}
    \centering
    \includegraphics[width=0.7\textwidth]{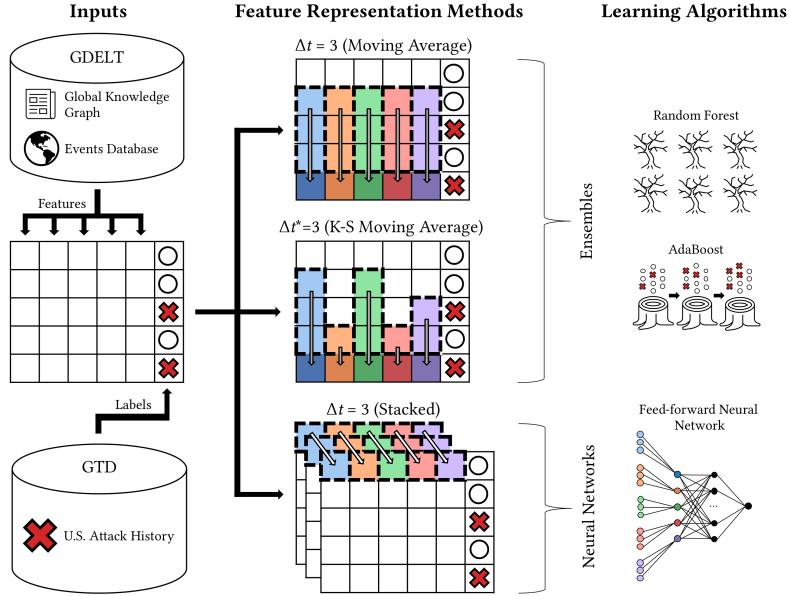}
    \caption{An overview of the feature extraction and representation procedure. Textual themes and CAMEO events from the Global Knowledge Graph and Events Database, respectively, were used to generate daily observations about each state. The ensemble methods learned from a moving average representation ($\Delta t$ and $\Delta t^*$) of the daily observations, and the neural network methods learned from a stacked representation. A red X indiciates an event (terrorist attack) while a white circle indicates a non-event (absence of terrorist attack), and each color represents a unique news feature.}
    \label{fig:obswindows}
\end{figure}

The Global Database of Events, Language, and Tone (GDELT) monitors worldwide print, broadcast, and online news in over 100 languages \cite{gdelt, leetaru2013gdelt}. It contains two primary sources of information: the Global Knowledge Graph (GKG) and the Event Database. The GKG comprises over 10 TB and grows at a rate of over 500,000 records every day. Each record represents publishing information, mentions of persons and locations, textual themes, and other metadata extracted from each published article, video, or other item of news. The Event Database utilizes the CAMEO event coding framework to identify and categorize events from news in the GKG. A CAMEO event consists of an actor---an individual, country, identity group, or some other sociopolitical entity---performing an action on another actor \cite{schrodt2012cameo}. Each action is classified under a taxonomy that includes categories such as ``acknowledge or claim responsibility'' (code 015) and ``reject request for military aid'' (code 1122). The Events Database records each event along with other extracted information, including the geographical location of the event and/or actors, political and religious affiliations for each actor, an estimate of the event’s political significance, and the average sentiment of news items associated with the event. Taken together, the GKG and Events Database represent GDELT's effort to construct a global representation of ``what's happening and how the world is feeling about it'' \cite{gdelt}. We utilized version 2 of the GKG and Events Database, which were both released on February 18, 2015.

We utilized data from both the GKG and Events Database to construct the feature matrix for each location, in a manner similar to Galla et al. \cite{galla2018predicting}. From the GKG, we considered 283 textual themes that are described in the GKG Category List \cite{gkgcat}. While other themes have been added to the GKG, the aforementioned are the best-documented have been clearly recorded since the release of GDELT version 2. For each theme, we extracted all the news records associated with that theme that also mention the given location. To compute the first set of 283 features, which we call \textbf{theme counts}, we simply counted the relevant records in the GKG. For the next 283 features, which we call \textbf{theme sentiments}, we computed the average sentiment score of these records for each date. From the Events Database, we grouped all records by the 148 ``base codes'' at the second level of the CAMEO taxonomy \cite{schrodt2012cameo}. The next set of 148 features, which we call \textbf{CAMEO counts}, is the count of news items associated with each event code for the given location, aggregated by publishing date. Our final set of 148 features, which we call \textbf{CAMEO sentiments} is the average sentiment score for all articles associated with each of the CAMEO codes, also aggregated by publishing date. In total, we considered 862 features, which collectively represent a location's sociopolitical climate on a given day.

\subsubsection{Label Extraction} \label{sec:labelcoll}
The Global Terrorism Database (GTD), created and maintained by the University of Maryland, is an open-source database containing information on over 190,000 terrorist events around the world from 1970 through 2018 \cite{lafree2007introducing}. The GTD defines a terrorist attack as ``the threatened or actual use of illegal force and violence by a non-state actor to attain a political, economic, religious, or social goal through fear, coercion, or intimidation'' \cite{start2019codebook}. According to its codebook, an event must meet the following requirements to be classified as a terrorist attack and recorded in the GTD:
\begin{enumerate}
    \item All three of the following criteria must be met:
    \begin{enumerate}
        \item The incident must be intentional.
        \item The incident must entail some level of violence or immediate threat of violence.
        \item The perpetrators of the incident must be sub-national actors.
    \end{enumerate}
    \item Two of the following three criteria must be met:
    \begin{enumerate}
        \item The act must be aimed at attaining a political, economic, religious, or social goal.
        \item  There must be evidence of an intention to coerce, intimidate, or convey some other message to a larger audience (or audiences) than the immediate victims.
        \item The action must be outside the context of legitimate warfare activities
    \end{enumerate}
\end{enumerate}

We considered all terrorist events that occurred in the United States on or after February 18, 2015---the release date of GDELT version 2. Of these 229 events, shown in Fig. \ref{fig:gtdhistory}, 83.8\% were successful, meaning that the attack produced some kind of tangible effect (such as an explosion), even if there were no casualties or significant consequences. 19.1\% of the attacks resulted in at least one fatality, and in 27.2\% at least one person was wounded. If multiple attacks are carried out in a continuous period of time or location, they are recorded as a single incident. However, even if the attacks are carried out in close proximity or by the same perpetrators, if either the time or location is discontinuous, multiple incidents are recorded. Therefore, from the 229 events, we preserved 208 unique location and date pairs as ground truth labels for the classification problem.

\subsection{Observation Windows}
Recall that our goal is to predict future terrorist attacks. However, the features and labels described above represent news activity and the occurrence of an attack, respectively, for the same day. In a real-world scenario, the news features for a given day would not be available until after the attack occurred. Additionally, the features only consider news activity for a single day; this precludes the discovery of any long-range temporal patterns. In other words, in its raw form the data could be used same-day terrorism detection, but not prediction. 

To address both of these shortcomings, we performed an additional preprocessing step to generate new feature representations for each date and location pair using what we call an \textbf{observation window} \cite{galla2018predicting}. This window, represented by the hyperparameter $\Delta t \geq 1$, is the number of prior days for which news features are observed when predicting an attack. As outlined in Fig. \ref{fig:obswindows}, we evaluated three distinct types of observation windows: a fixed-length moving average, a variable-length moving average, and a stacked representation. Each approach is detailed below.

\subsubsection{Fixed-length Moving Average} In the simplest case, for each feature $X_{*,j}$, we considered the unweighted mean of the previous $\Delta t$ days, i.e.
\begin{equation} \label{eq:deltat}
MA(X_{*,j}, \Delta t) = \left[ \frac{\sum_{k=1}^{\Delta t}X_{i-k,j}}{\Delta t} : i \leq n \right].
\end{equation}

Note that when $\Delta t = 1$, the representation for a given location and date is simply the news features as observed from the previous day. Increasing $\Delta t$ allows us to observe a greater number of prior days and has the additional benefit of smoothing noise. However, it also blurs the distinctions between days and causes each feature to no longer be independently distributed; we address this problem in Section \ref{sec:expsetup}.

\subsubsection{Variable-length K-S Moving Average} A significant limitation of the fixed-length approach is that it assumes a static value of $\Delta t$ for all features. We addressed this by implementing a more flexible version of the observation window, which chooses a preferred representation for each feature according to a two-sample Kolmogorov-Smirnov (K-S) test \cite{hodges1958significance}. This procedure is detailed in Algorithm \ref{alg:deltatstar}. In summary, we first specify a maximum observation window $\Delta t^*$. Next, for each feature we compute its moving moving average (Equation \ref{eq:deltat}) for each $t \leq \Delta t^*$. Then for each feature and value of $t$ we perform a K-S test, such that the first sample is the set of observations for that feature for non-events and the second sample is the set of observations for events. Finally, we choose the representation for each feature (i.e., value of $t$) that maximizes the distance between the empirical distribution functions of events and nonevents (i.e., returns the minimum $p$-value).

\begin{algorithm}
\caption{K-S Moving Average}
\label{alg:deltatstar}
\begin{algorithmic}[1]
    \Require A feature matrix $X \in \mathbb{R}^{n \times m}$, a label vector $\vec{y} = \{0,1\}^n$, a maximum observation window $\Delta t^* \geq 1$
    \Ensure A processed feature matrix $X$
    \For{$j \gets 1$ to $m$}
        \State $p_{min} \gets 1.0$ \Comment{Store the minimum p-value}
        \State $a \gets X_{*,j}$ \Comment{Store the original observations for feature $j$}
        \For{$t \gets 1$ to $\Delta t^*$}
            \State $a' \gets MA(a, t)$ \Comment{Compute the new moving average}
            \State $p \gets KS(\{a'_i : y_{i} = 0\}, \{a'_i : y_{i} = 1\}$) \Comment{Compute the new p-value using the K-S test}
            \If{$p < p_{min}$}
                \State $p_{min} \gets p$ \Comment{Store the new minimum}
                \State $X_{*,j} \gets a'$ \Comment{Replace the values for feature $j$}
            \EndIf
        \EndFor
    \EndFor
    \State \Return $X$
 \end{algorithmic}
\end{algorithm}

\subsubsection{Stacked Representation} \label{sec:stacked}
Both moving average approaches described previously are unweighted and thus assume each of the values within the observation window are equally important. While this may be suitable for simpler learning algorithms, neural networks are well-equipped to discover more complex patterns by generating their own (hidden) representations. Therefore, when serving input to neural networks we do not precompute representations using either of the moving average approaches described above; instead, we represent each instance as stacked feature vectors from the previous $\Delta t$ days, i.e. $X_{i-\Delta t:i-1}$ for some arbitrary day $i$. By learning a mapping from these stacked inputs to a hidden representation, the neural network is essentially learning its own observation window. We discuss this in more detail in the following section.

\subsection{Machine Learning Models} \label{sec:models}
Recall that our task is to learn a function $f : x \xrightarrow{} \{0,1\}$ for some input $x$. Toward this end we explored two families of machine learning algorithms, ensembles and neural networks, which we detail below.
\subsubsection{Ensembles}
An ensemble is a model that is comprised of multiple classifiers. When individual classifiers are independent and diverse, the ensemble can mitigate the weaknesses and limitations of a single classifier. This characteristic has enabled ensembles to provide state-of-the-art performance on a number of machine learning tasks  \cite{sagi2018ensemble}. We specifically utilized the following two ensembles:
\begin{enumerate}
    \item Random Forest: a collection of unpruned decision trees, each grown from a random and uniformly sampled subspace (of features), that uses a weighted voting mechanism to classify each instance \cite{breiman2001random}. The Random Forest has found success in a range of machine learning tasks and established itself as a strong general-purpose ensemble \cite{sagi2018ensemble}.
    \item AdaBoost: a collection of decision stumps that are each trained on a single bootstrap (a small sample drawn with replacement from the training set). AdaBoost's key innovation is that it iteratively updates the sampling distribution, increasing the probability of selection in the next bootstrap for instances that were misclassified by the previous decision stump \cite{freund1997decision}. This procedure intuitively pushes the ensemble to ``focus'' on more difficult instances; however, it also makes AdaBoost very sensitive to noise.
\end{enumerate}
\subsubsection{Neural Networks}
A neural network propagates an instance through layers of nodes, or neurons, to compute its label. Training a neural network is the process of learning the set of edge weights that minimizes error on the training data with respect to a loss function. The power of neural networks lies in their use of various configurations of hidden (intermediate) layers to generate new representations of the input. As such, neural networks have become state-of-the-art in many complex learning tasks---especially those that involve learning nonlinear functions \cite{sagi2018ensemble}. A neural network is a function that classifies an input $x$ according to the following:
\begin{equation}
\label{eq:nn}
    f(x) = \sigma(W h(x) + \vec{b}),
\end{equation}

where $W$ is a weight matrix and $\vec{b}$ is a bias vector that are both learned via backpropagation; $\sigma$ is the sigmoid activation function, and $h(x)$ is the learned hidden representation of the raw input $x$. What distinguishes neural networks is how they compute $h(x)$. For brevity and in this section only, we use $X_{i-\Delta t:i-1}$ to represent the matrix form of the stacked input for an arbitrary day $i$ and observation window $\Delta t > 0$ , as introduced in Section \ref{sec:stacked}. Given this definition of $X'$, we utilized the following neural network functions:
\begin{enumerate}
\item Feedforward Neural Network (FFNN): a basic architecture in which inputs are propagated strictly forward from the input layer, through a series of $L$ hidden layer(s), and finally to the output layer. We utilize two distinct FFNN architectures:
\begin{enumerate}
    \item One hidden layer ($L = 1$, not pictured): computes $h(X')$ via a single dense layer, i.e.
    \begin{equation}
        h(X') = ReLU \big( W_0 \left[ X'_{*,1}, ..., X'_{*, m} \right] + \vec{b_0} \big),
    \end{equation}
    where $ReLU$ is a rectified linear unit, $W_0 \in \mathbb{R}^{m \times k}$ is a learned weight matrix with $k$ (a hyperparameter) neurons in the hidden layer, $[ X'_{*,1}, ..., X'_{*, m}]$ represents the vector concatenation of all observations of all features in $X'$, and $\vec{b_0} \in \mathbb{R}^k$ is a learned bias vector for the hidden layer.
    
    \item Two hidden layers ($L = 2$, pictured in Fig. \ref{fig:obswindows}): computes a hidden representation for each feature independently, then uses each feature's new representation to compute a joint hidden representation via a dense layer:
    \begin{equation}
        h(X') = ReLU \big(W_0 \left[ g (X'_{*,1}), ..., g(X'_{*,m}) \right] + \vec{b_0} \big),
    \end{equation}
    \begin{equation}
        g(X'_{*,j}) = ReLU \big(W_j X'_{*,j} + \vec{b_j} \big).
    \end{equation}
    Here $W_j$ is the weight matrix and $b_j$ the bias vector for the $j^{th}$ feature. 
\end{enumerate}
\item Long Short-term Memory (LSTM): an architecture in which inputs are propagated through a series of recurrent layers in order to learn sequential dependencies \cite{hochreiter1997long}. LSTMs have found success in time series classification, machine translation, and a number of other tasks in which the input can be modeled as a sequence. We use a standard sequential LSTM to compute a hidden representation for each day $t < \Delta t$ according to:
\begin{equation}
    h_t(X') = tanh (W_t \left[\vec{x'_t}, h_{t-1}(X'_{1:t-1}) \right] + \vec{b_t}),
\end{equation}
where $tanh$ is the standard $tanh$ activation function and $\vec{x'_t}$ is the feature vector for day $t$ in $X'$. 
\end{enumerate}

We experimentally evaluated several other state-of-the-art neural network architectures, including models based on attention \cite{vaswani2017attention} and convolutions, which are often used in time series classification and have the benefit of being translation invariant \cite{fawaz2019deep}. However, we found that these additional architectures performed empirically worse than both the FFNN and LSTM, so we do not consider them in the rest of this study.

\subsection{Addressing Class Imbalance}
When trained on imbalanced data, the predictions of machine learning models can be biased toward the majority class. We utilized the following techniques to address this problem:
\subsubsection{Ensembles: SMOTE} Synthetic minority oversampling technique (SMOTE) generates artificial examples of the minority class \cite{chawla2002smote} in order to balance the training set. We utilized SMOTE in all the ensembles by oversampling the minority class (events) such that both classes had equal representation.

\subsubsection{Neural Networks: Weighted Cross-Entropy} Because backpropagation trains a neural network by computing the gradient with respect to a loss function, a useful technique in imbalanced settings is to weigh more heavily the loss contributions from the minority class. Toward this end we utilized a weighted version of cross-entropy (log loss). Consider the standard definition of binary cross-entropy, i.e.

\begin{equation}
    H(\vec{y}) = -\frac{1}{n} \sum_{i = 0}^{n} \Big(y_i  log_2 \big( p(y_i) \big) + (1 - y_i) log_2 \big( 1 - p(y_i) \big) \Big),
\end{equation}

where $p(y_i)$ is the probability output by the model that example $y_i = 1$ (is an event). We added a term $\alpha = \frac{\sum_{i=0}^{n}y_i}{n}$ to weigh the contribution of each training example by the inverse of its class representation:

\begin{equation}
    H_w(\vec{y}) = -\frac{1}{n} \sum_{i = 0}^{n} \Big( (1-\alpha) y_i  log_2 \big( p(y_i) \big) + \alpha (1 - y_i) log_2 \big( 1 - p(y_i) \big) \Big).
\end{equation}
This weighting allows for the set of events and the set of nonevents to each contribute half to the potential loss. While we also evaluated SMOTE and other functions like focal loss \cite{lin2017focal} and class-balanced cross-entropy \cite{cui2019class}, we found that weighted cross-entropy provided the best performance for the neural networks.

\subsection{Experimental Setup} \label{sec:expsetup}
All experiments utilized Python 3.7.3 in conjunction with sklearn 0.22.2 (ensemble methods) and Keras 2.2.4 with TensorFlow 2.0 (neural networks). For the Random Forest and AdaBoost we used 3,000 estimators and Gini index as the splitting criterion for each tree. We trained all neural networks for 100 epochs using stochastic gradient descent with a learning rate of $1e^{-4}$, a decay parameter of $1e^{-6}$, a batch size of 32, and Nesterov momentum. All hidden layers (except the first hidden layer in the two-layer FFNN architecture; see Section \ref{sec:stacked}) contained 8,000 neurons, and each hidden layer in the LSTM model contained 1,024 neurons. For the neural network models we used min-max normalization to scale the input. Beyond computing the observation windows, No further preprocessing was required for the ensembles. We tested the fixed-length moving average approach with the ensemble methods and the stacked representation with the neural networks. However, due to the success of the K-S moving average ($\Delta t*$) with the ensembles, we also tested this representational method with the neural networks.

We evaluated each model using 5-fold cross-validation. This was straightforward for the neural network models and when $\Delta t = 1$, since each observation is independent. However, when using either moving average approach we included additional considerations in order to maintain proper separation between the training and testing sets:
\begin{enumerate}
    \item When testing the variable-length moving average ($\Delta t^*$), we used only the training data to compute the optimal observation windows for each feature. We then computed moving averages on the testing set using the same window lengths that were learned from the training set.
    \item When testing either type of moving average, we removed any examples from the training set whose values were dependent on examples in the testing set. For example, consider the case where $\Delta t = 14$ and the testing fold is the 282 days from Mar. 25 through Dec. 31, 2018. The feature values for Mar. 25 will be highly correlated with the values for Mar. 11-24, which are in the training set, since their moving averages are computed using at least one of the same observations. We resolved this by dropping Mar. 11-24 from the training set. This approach handicaps the moving average models (since they are provided with fewer training examples), but ensures that the testing sets are consistent across all models.
\end{enumerate}

Finally, we measured model performance on each testing set using area under the receiver operating characteristic curve (AUROC). An ROC curve plots a model's true positive rate against its false positive rate at various discrimination thresholds, and is thus a useful measure of classification performance in an imbalanced setting. An AUROC of 1.0 indicates a perfect classifier, while a random guesser converges to 0.5. We repeated each 5-fold cross-validation experiment 10 times and report both the mean and standard deviation AUROC values for each model.

%% file: results.tex
\section{Results and Discussion} \label{sec:results}

\begin{table*}[]
    \centering
    \begin{tabular}{cc|c|ccccc}
         & \multirow{2}{*}{\textbf{Model}} & \textbf{Observation} & \multicolumn{5}{|c}{\textbf{AUROC (5-fold cross validation)}} \\ 
         
         & & \textbf{Window} & \textbf{NY} & \textbf{California} & \textbf{Texas} & \textbf{Florida} & \textbf{Washington} \\ \cline{2-8}
         
         & Random Baseline & --- & $.500 \pm .000$ & $.500 \pm .000$ & $.500 \pm .000$ & $.500 \pm .000$ & $.500 \pm .000$ \\ \cline{2-8}
         
         \multirow{6}{*}{\rotatebox[origin=c]{90}{Ensembles}} & \multirow{3}{*}{Random Forest} & $\Delta t=1$ & $.604 \pm .118$ & \underline{$.504 \pm .065$} & \underline{$.721 \pm .184$} & $.591 \pm .140 $  & $.500 \pm .248$ \\
         
         & & $\Delta t=14$ & $.631 \pm .093 $ & $.390 \pm .133$ & $.530 \pm .166$ & $.623 \pm .120$ & $.591 \pm .261$ \\
         
         & & $\Delta t^{*}=14$ & \underline{$\boldsymbol{.685 \pm .057}$} & $.466 \pm .060$ & $.682 \pm .196$ & \underline{$.685 \pm .101$} & \underline{$\boldsymbol{.667 \pm .214}$} \\
         
         & \multirow{3}{*}{AdaBoost} & $\Delta t=1$ & $.623 \pm .079$ & $.436 \pm .110$ & $.574 \pm .117$ & $.459 \pm .093$ & $.376 \pm .122$\\
         
         & & $\Delta t=14$ & $.500 \pm .243$ & $.360 \pm .189$ & $.401 \pm .136$ & $.421 \pm .126$ & $.538 \pm .253$ \\
         
         & & $\Delta t^{*}=14$ & $.668 \pm .073$ & $.463 \pm .135$ & $.589 \pm .136$ & $.472 \pm .123$ & $.544 \pm .171$\\ \cline{2-8}
        \multirow{5}{*}{\rotatebox[origin=c]{90}{Neural Nets}} & \multirow{4}{*}{FFNN} & $\Delta t=1, L=1$ & $.393 \pm .065$ & $.584 \pm .100$ & $.454 \pm .199$ & $.420 \pm .099$ & $.409 \pm .149$ \\
         
         & & $\Delta t=7$, $L=1$ & $.414 \pm .110$ & $.672 \pm .095$ & $.650 \pm .118$ & $.667 \pm .098$ & $.615 \pm .252$ \\
         & & $\Delta t=7$, $L=2$ & \underline{$.428 \pm .095$} & \underline{$\boldsymbol{.680 \pm .073}$} & $.528 \pm .127$ & $.362 \pm .144$ & \underline{$.648 \pm .201$} \\
         & & $\Delta t^{*}=7$, $L=1$ & $.479 \pm .219$ & $.568 \pm .077$ & $.591 \pm .175$ & $.552 \pm .039$ & $.546 \pm .229$ \\
         & \multirow{1}{*}{LSTM} & $\Delta t=7$ & $.374 \pm .078$ & $.582 \pm .153$ & \underline{$\boldsymbol{.725 \pm .212}$} & \underline{$\boldsymbol{.696 \pm .168}$} & $.563 \pm .270$ \\ \cline{2-8} 
          
    \end{tabular}
    \caption{Summary of classification results. The reported values are the means and
    standard deviations as calculated by 5-fold cross validation. Underlined values indicate the best-performing model in each group (ensembles and neural networks) and bold values indicate the best-performing model overall.}
    \label{tab:resultsmain}
\end{table*}

Table \ref{tab:resultsmain} summarizes the classification results for observation windows of 1 and 14 days for both ensembles (Random Forest and AdaBoost), and 1 and 7 days for the neural networks (FFNN and LSTM). While some other observation windows produced better results for individual states, those shown in Table \ref{tab:resultsmain} produced at least one model that was strong across several states, and thus seemed the most useful. The standard deviation for each complete 5-fold cross-validation experiment was small ($< 0.025$ in all cases), but the inter-fold standard deviation was high in many cases. For example, on New York, the Random Forest ($\Delta t* = 14$) model consistenly performed well on the first testing fold ($.726 \pm .036$) but struggled on the fifth fold ($.615 \pm .036$). This is not surprising, given that individual events are difficult to predict and that the small number of events in our data set means it is less likely that each training and testing fold will reflect the true distribution. In spite of this, we found that multiple models achieved results that are noteworthy and signifcantly better than random. In particular, the random forest with $\Delta t^*=14$ achieved an AUROC $\geq .667$ for every state except California, on which all the ensembles performed poorly. The neural networks performed better on California, but this was exchanged for poor performance on New York. The neural network that performed best on average was the FFNN with $\Delta t=7, L = 1$, which achieved an AUROC $\geq .650$ for three of five states. The other neural networks performed well on two states---California and Washington for the FFNN with $\Delta t^*=7, L = 2$, and Texas and Florida for the LSTM---but poorly on the others. Overall we consider the Random Forest ($\Delta t^*$) to be the strongest model, followed by the FFNN ($L=1$), since it produced the best results when averaged across all states. As such, we focus the remainder of our discussion on the performance of these two models and refer to them as $RF$ and $FFNN$, respectively.

\subsection{Observation Windows}
\begin{figure}
    \centering
    \includegraphics[width=0.8\textwidth]{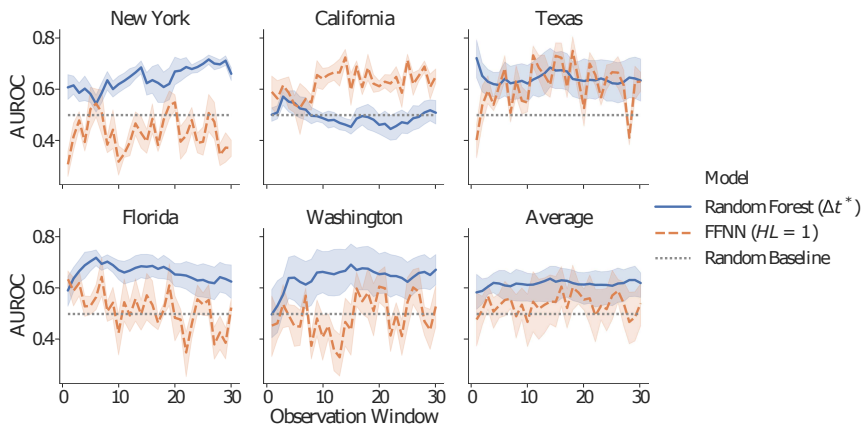}
    \caption{Classification results for different observation windows on RF and FFNN show that the optimal window varies between states. The x-axis represents the observation window (number of previous days considered in the feature representation for each instance), and the y-axis represents the average AUROC computed using 5-fold cross validation. The shaded areas represent standard deviation across all testing folds.}
    \label{fig:deltas}
\end{figure}

Figure \ref{fig:deltas} depicts model performance as a function of observation window (each tick on the x-axis is a separate model) for RF and FFNN. We note that the trend of RF's performance is relatively stable with respect to increasing observation windows. FFNN, on the other hand, is relatively sensitive to small changes to the observation window. This instability is likely attributable to noise, and suggests that FFNN is much more prone to overfitting than RF. From this result we additionally note that the trend and optimal window varies between states. While this suggests that a predictive model may need to be tuned separately for each state, it is also possible that some of this variation is due to noise and/or only having a small number of events per state.

\subsection{Temporal Locality of Attacks}

\begin{figure}
    \begin{subfigure}{\textwidth}
        \centering
        \includegraphics[width=0.65\textwidth]{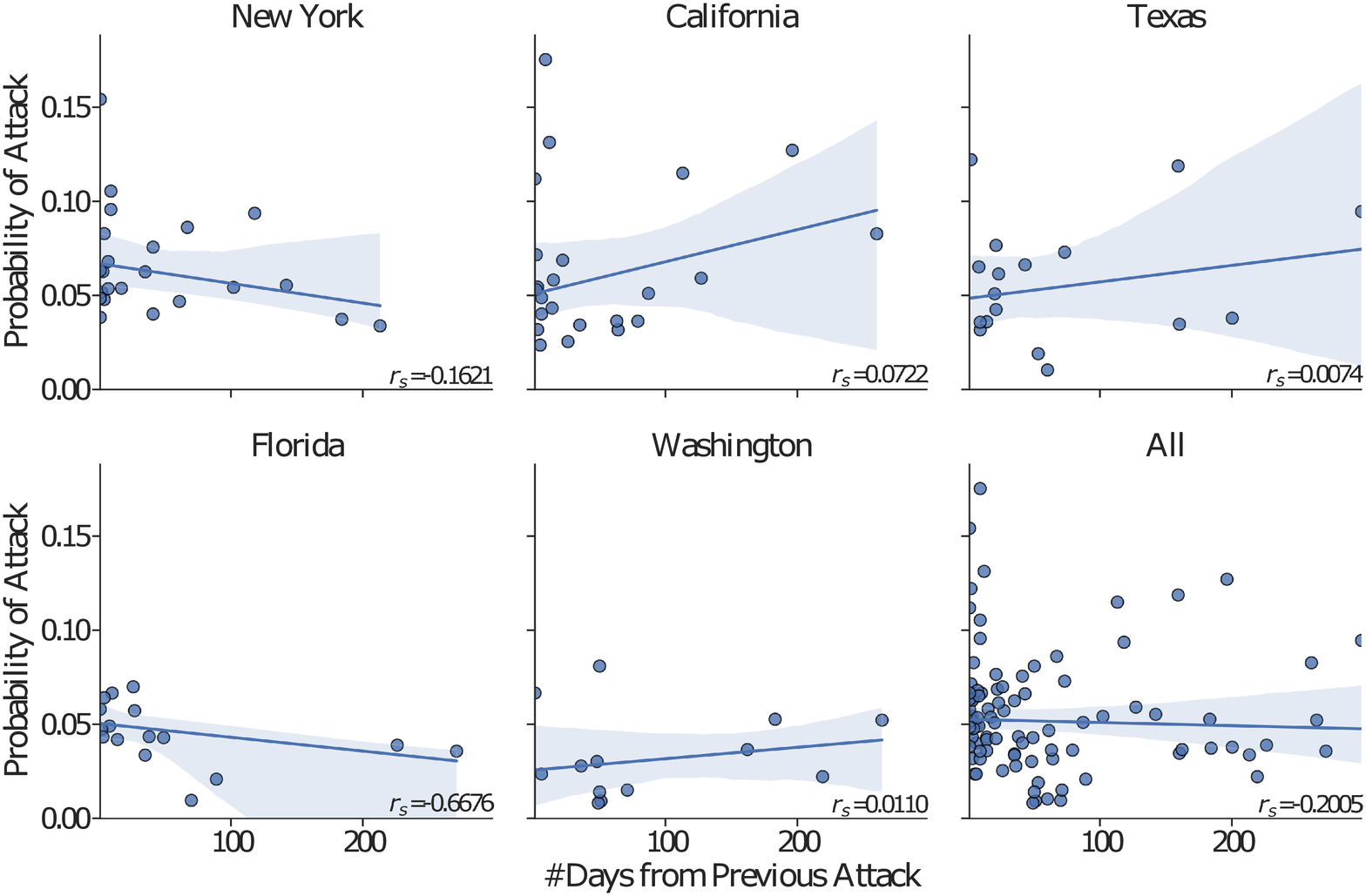}
        \caption{RF results.}
        \label{fig:rftimecorr}
    \end{subfigure}
        \begin{subfigure}{\textwidth}
        \centering
        \includegraphics[width=0.65\textwidth]{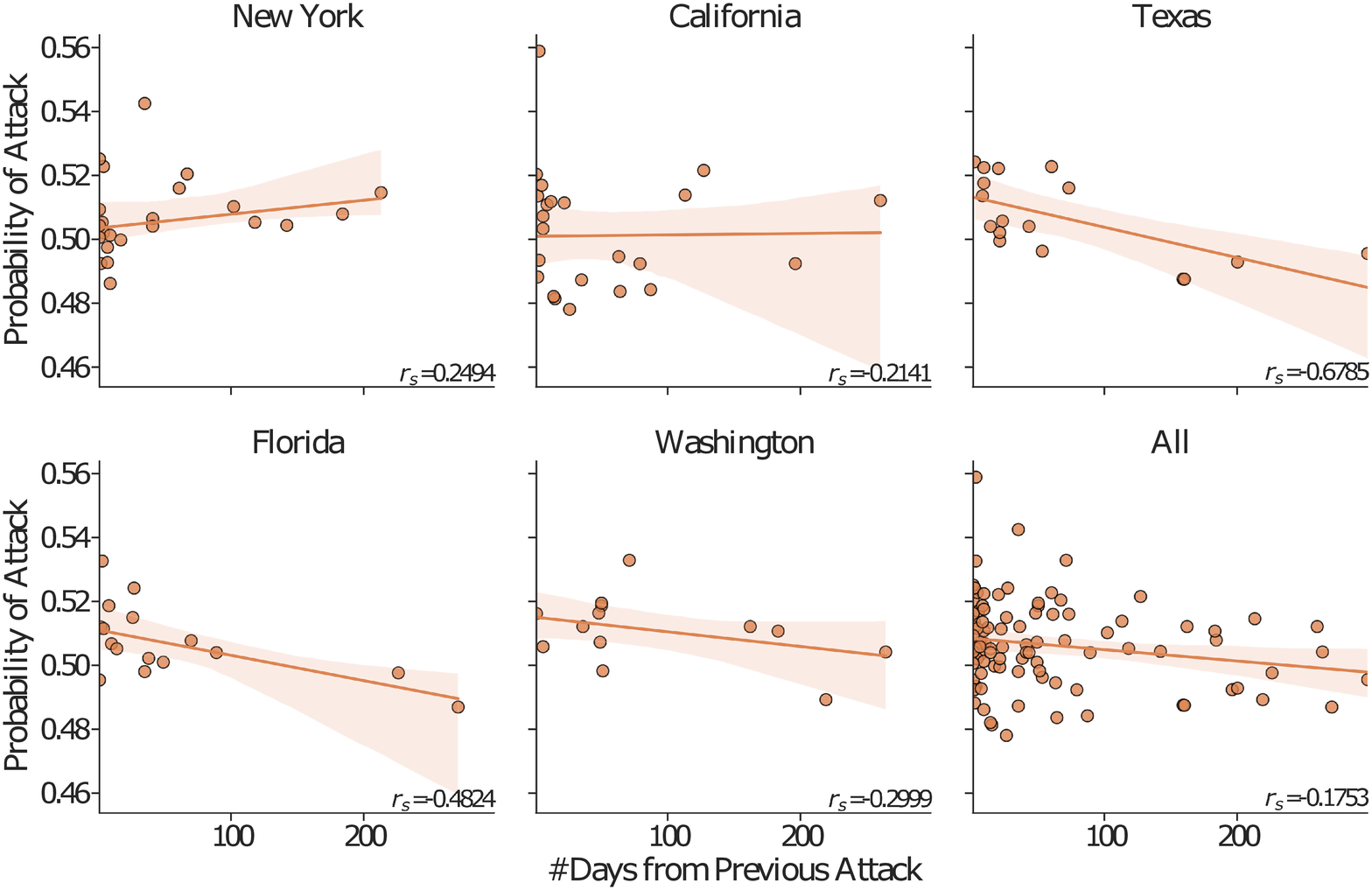}
        \caption{FFNN results.}
        \label{fig:fftimecorr}
    \end{subfigure}
    \caption{The predicted probabilities for each attack, measured as a function of distance in time from the previous attack in that state, show that RF's predictions are relatively stable with respect to temporal locality. The shaded region is the confidence interval for the regression line, and $r_s$ is Spearman's rank correlation coefficient. The difference in scale in the y-axis between RF and FFNN is caused by differences in the training process for each model.}
    \label{fig:timecorr}
\end{figure}

Some of the attacks recorded in the GTD are clustered in time and space. For example, in 2016 attacks were carried out in New York on August 9 (anti-Semitic extremists detonated explosives outside the houses of two Rabbis in New York City), August 10 (an arsonist targeted a private residence in Endicott), and August 13 (a man shot and killed an Imam and his assistant outside a mosque in New York City). Additionally, there are a small number of attacks that spanned a number of days and/or targets, such as in October 2018, when pro-Trump extremists sent pipe bombs to five different targets on four different dates. To test whether temporal locality had biased the model outputs, we recorded for each attack the predicted probability ($P(y=1)$) and the number of days that had elapsed since the previous attack in that state. Figures \ref{fig:rftimecorr} and \ref{fig:fftimecorr} visualize this distribution for RF and FFNN, respectively, along with a regression line for each state. These results, along with the correlation coefficient from Spearman's rank test ($r_s$), indicate that RF still performs well on many of the attacks that are not close together in time, which is an important characteristic of a robust model. FFNN performed more poorly on attacks that were farther apart in time, which likely contributed to its weaker overall performance.

\subsection{Number of Attacks in Training and Testing}
\begin{figure}
    \begin{subfigure}{0.265\textwidth}
        \centering
        \includegraphics[width=\textwidth]{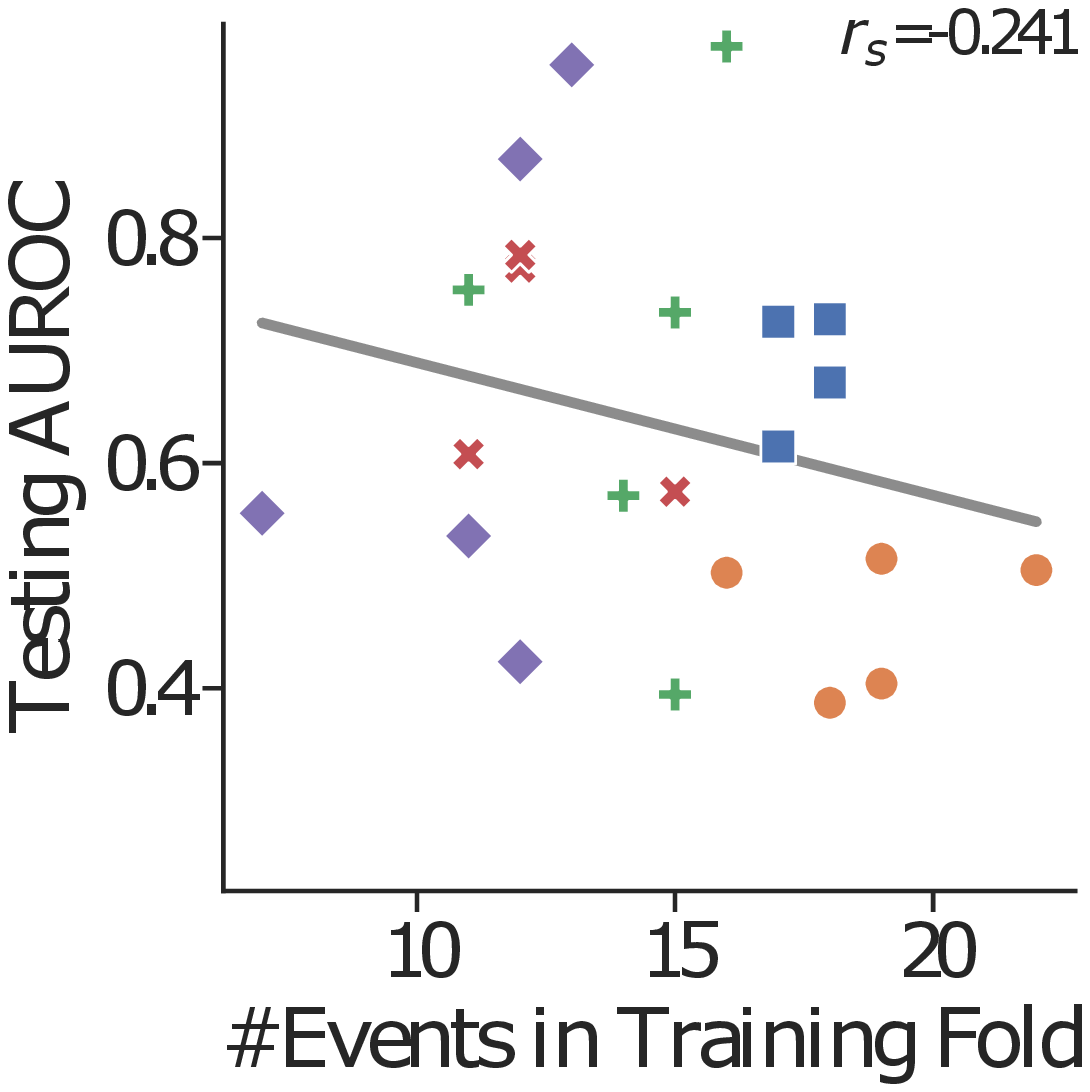}
        \caption{RF results.}
        \label{fig:rftraineventsize}
    \end{subfigure}
    \begin{subfigure}{0.372\textwidth}
        \centering
        \includegraphics[width=\textwidth]{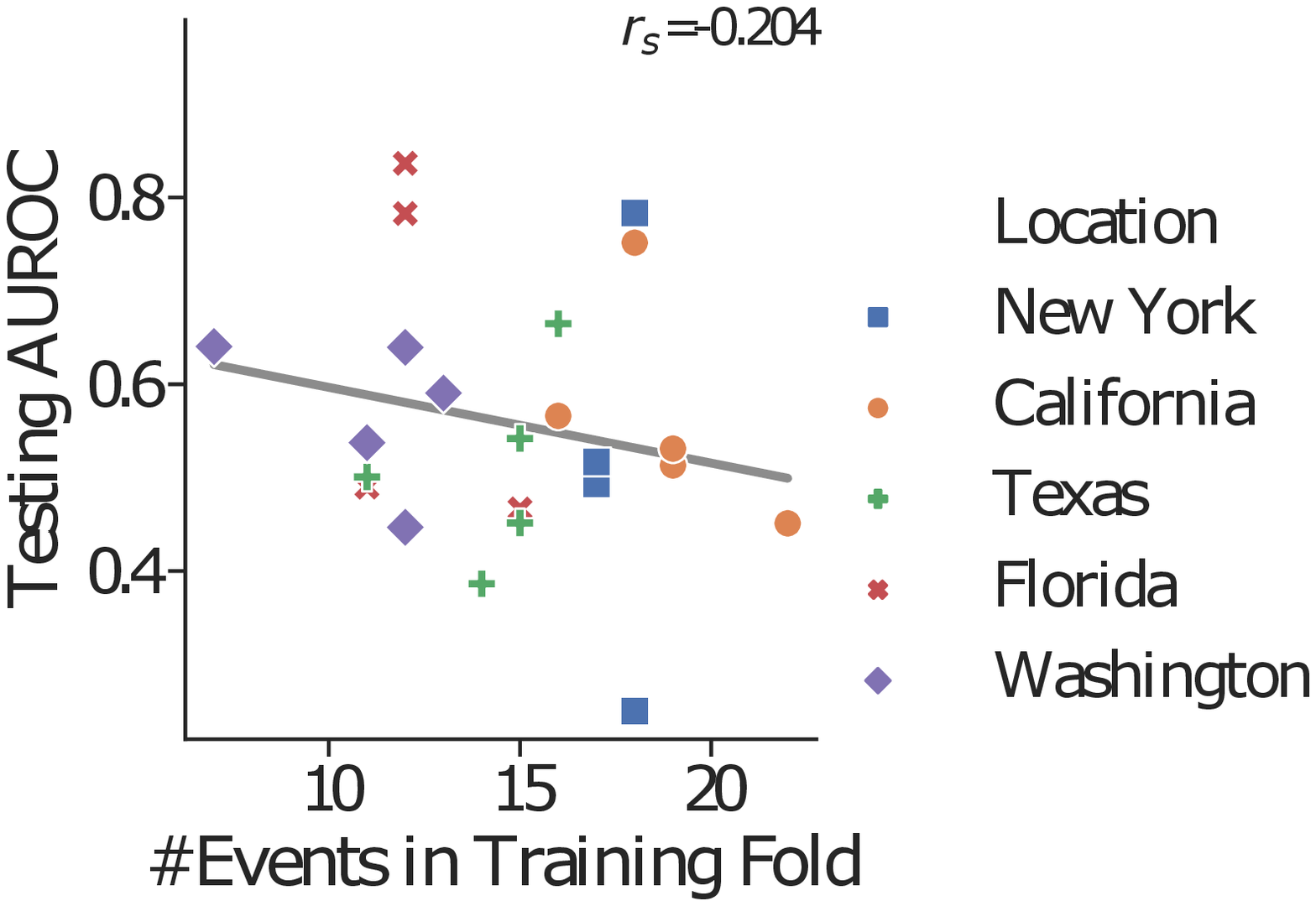}
        \caption{FFNN results.}
        \label{fig:fftraineventsize}
    \end{subfigure}
    \caption{The AUROC of each test fold, plotted as a function of the number of attacks in the corresponding training fold, shows that having more attacks in the training set does not necessarily increase performance on the test set. The grey lines are regression lines, and $r_s$ is Spearman's rank correlation coefficient.}
    \label{fig:rfeventsize}
\end{figure}
The fact that our data set is both small and highly imbalanced is almost certainly a limiting factor in model performance. Additionally, the temporal nature of our data limits our ability to use stratified cross validation, so the events are not distributed evenly across training and testing folds. In Figures \ref{fig:rftraineventsize} and \ref{fig:fftraineventsize}, the AUROC of each testing fold is shown as a function of the number of events in the corresponding training fold for RF and FFNN, respectively. When computed across all states, the line of best fit and Spearman's rank coefficient ($r_s$) for these results surprisingly suggest a negative correlation---i.e., that more training and testing examples is correlated with worse performance. However, in the case of RF, we can see that the folds from California---on which RF's predictions were close to random guessing---contribute significantly to this result since they contain many events. Without California, $r_s$ increases from $-0.241$ to $0.136$ for RF---the positive correlation we would expect. However, for both RF and FFNN it seems that the lack of events or data is not the only limiting factor. It is likely that data quality and/or noise are additional barriers, ones that are more pronounced given the small number of events.

\subsection{Feature Types}

\begin{figure}
    \centering
    \includegraphics[width=0.8\textwidth]{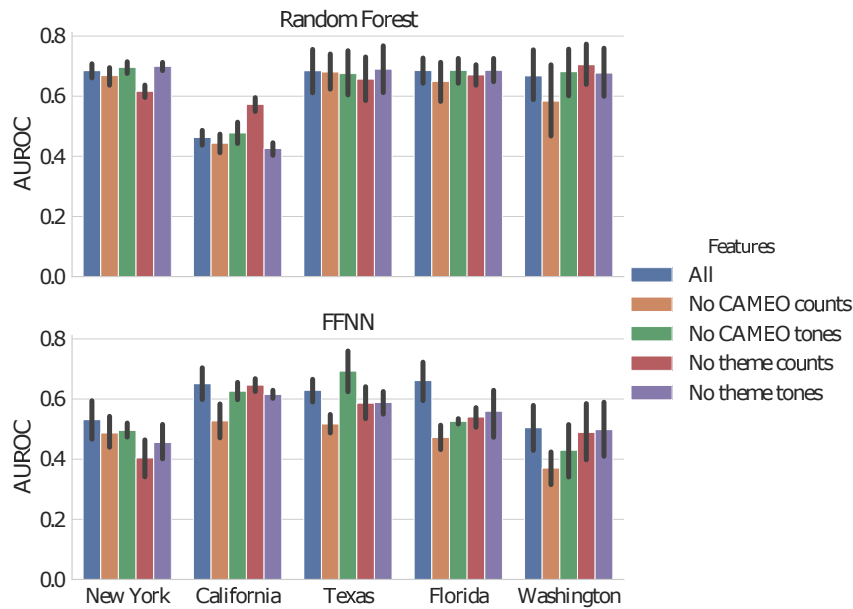}
    \caption{Classification results for each state using the baseline model along with four additional models, each of which was trained without one of the major feature groups described in Section \ref{sec:featureextraction}. Error bars represent standard deviations between folds in each 5-fold cross validation experiment.}
    \label{fig:featurex}
\end{figure}

Understanding the role of each group of features can help us mitigate noise and more deeply understand how our features contribute to successful (or unsuccessful) predictions. Toward this end, we retrained RF and FFNN using different combinations of feature groups. Figure \ref{fig:featurex} shows the baseline classification results on each state along with results from four additional feature-filtered models, and is the basis for the following observations:
\begin{enumerate}
    \item In general, the importance of each feature group varied between states. This affirms the need to consider a unique model for each state.
    \item The only feature group that improved performance in all cases was CAMEO counts. Further, in most states the model without CAMEO counts performed the worst. This implies that CAMEO counts are, on average, the most potent feature group.
    \item In some cases, dropping a feature group improved model performance; for example, RF's predictions on California. This is further evidence of the impact of noise on classification results, and creates an important point of investigation for any future work that would seek to optimize predictions for a particular state or location.
\end{enumerate}

\subsection{Characteristics of Attacks}

\begin{figure}
    \centering
    \begin{subfigure}{0.257\textwidth}
        \includegraphics[width=\textwidth]{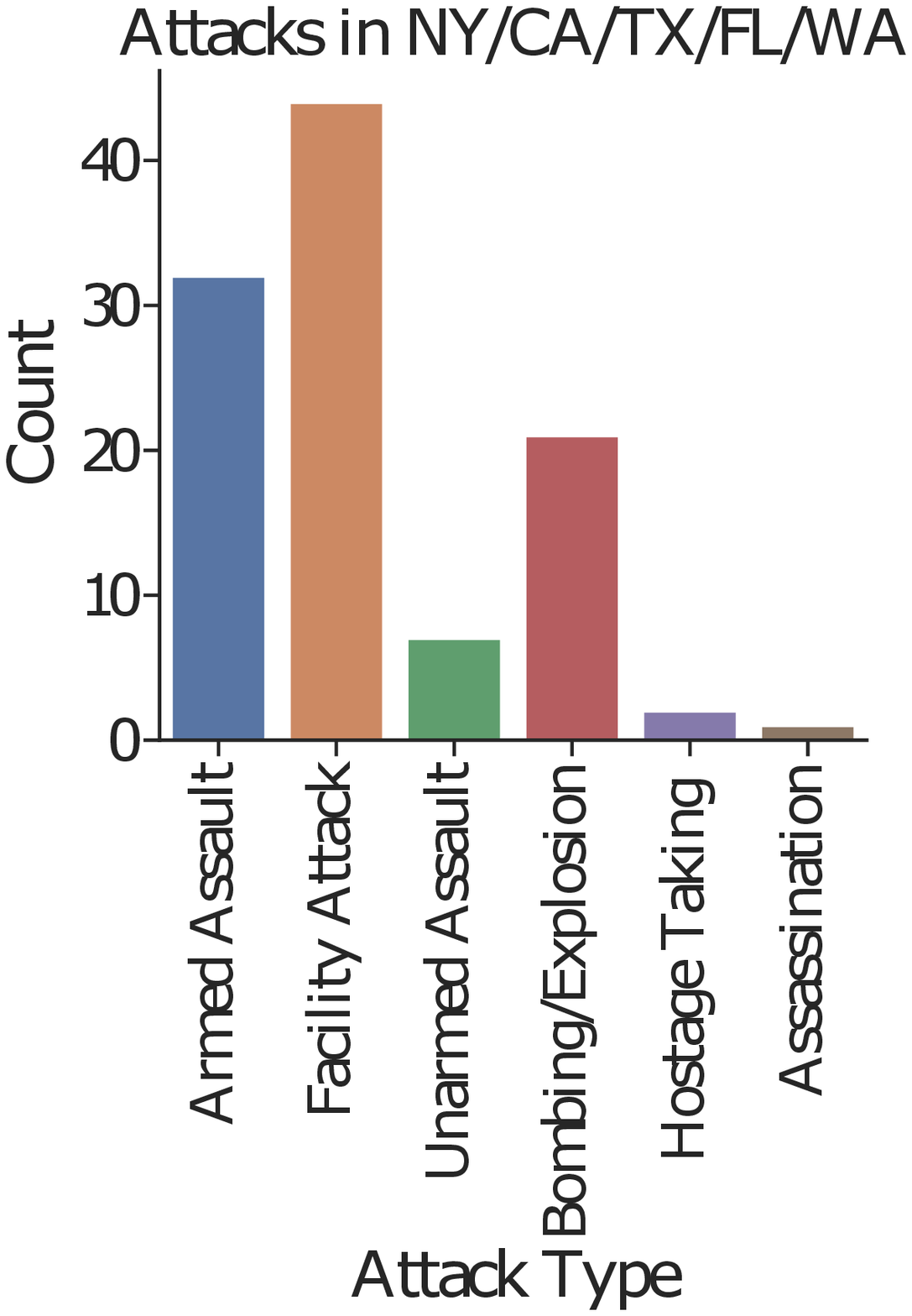}
        \caption{Distribution of attacks by type as recorded in the GTD.}
        \label{fig:attacktypecounts}
    \end{subfigure}
    \hspace{0.05\textwidth}
    \begin{subfigure}{0.25\textwidth}
        \centering
        \includegraphics[width=\textwidth]{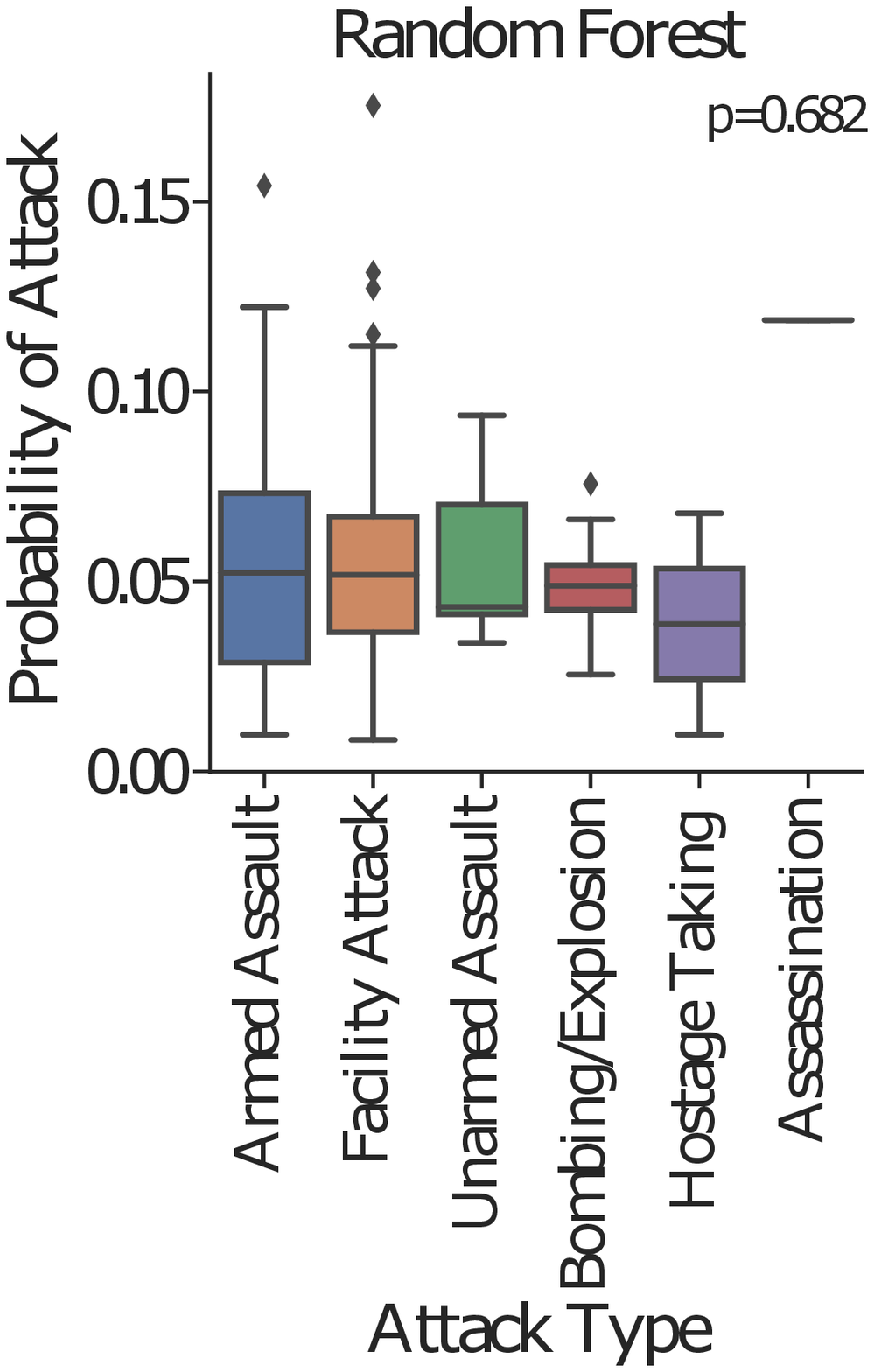}
        \caption{Distribution of RF's predicted probabilities ($P(y=1)$) for each attack type.}
        \label{fig:attacktyperf}
    \end{subfigure}
    \hspace{0.05\textwidth}
    \begin{subfigure}{0.25\textwidth}
        \includegraphics[width=\textwidth]{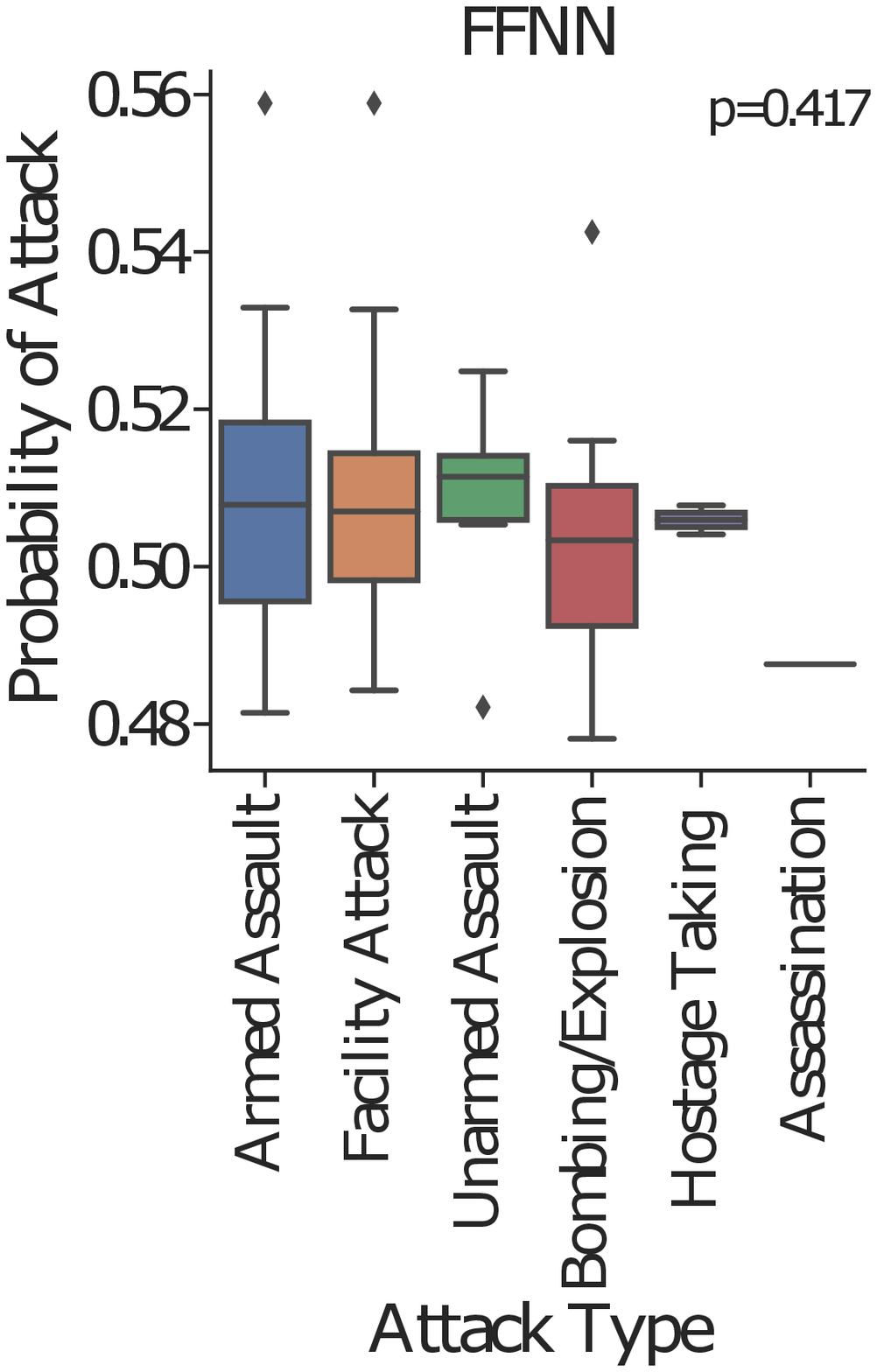}
        \caption{Distribution of FFNN's predicted probabilities ($P(y=1)$) for each attack type.}
        \label{fig:attacktypeffnn}
    \end{subfigure}
    \caption{Distribution of event count and predicted probabilities by attack type. The difference in scale in the y-axis between RF and FFNN is caused by differences in the training process for each model. In the box plots, each box represents observations within 1 standard deviation of the mean, which is represented by the horizontal line. Error bars represent observations within 3 standard deviations of the mean, while individual points represent outliers.}
    \label{fig:attacktypes}
\end{figure}

Terrorist attacks vary widely with respect to responsible party, target, method and other properties. Some types of attacks may be easier or harder to predict, or even characterized by different feature distributions. We used a Kruskal-Wallis H-test to test whether the median predictions from RF or FFNN differed significantly by each of the GTD's prominent categorical features: attack type (e.g., armed assault, bombing), primary weapon type (e.g., firearms, explosives), target type (e.g., military, religious institutions), and responsible group. As Figure \ref{fig:attacktypes} shows, each test produced a $p$-value $> 0.1$, so we cannot conclude that there is a correlation between the characteristics of an attack and prediction success.


\subsection{Prediction Windows}

\begin{figure}
    \centering
    \includegraphics[width=.8\textwidth]{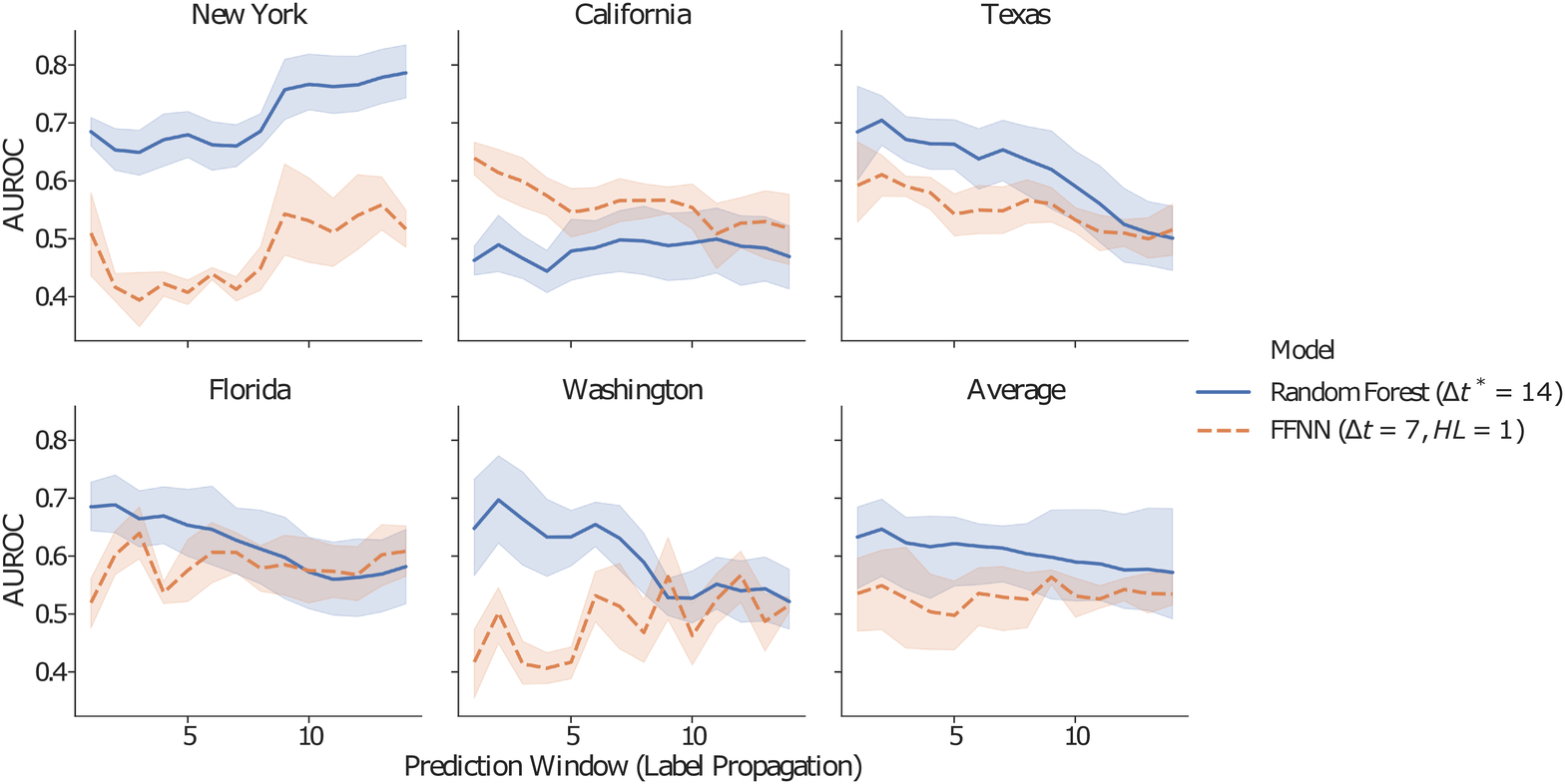}
    \includegraphics[width=.8\textwidth]{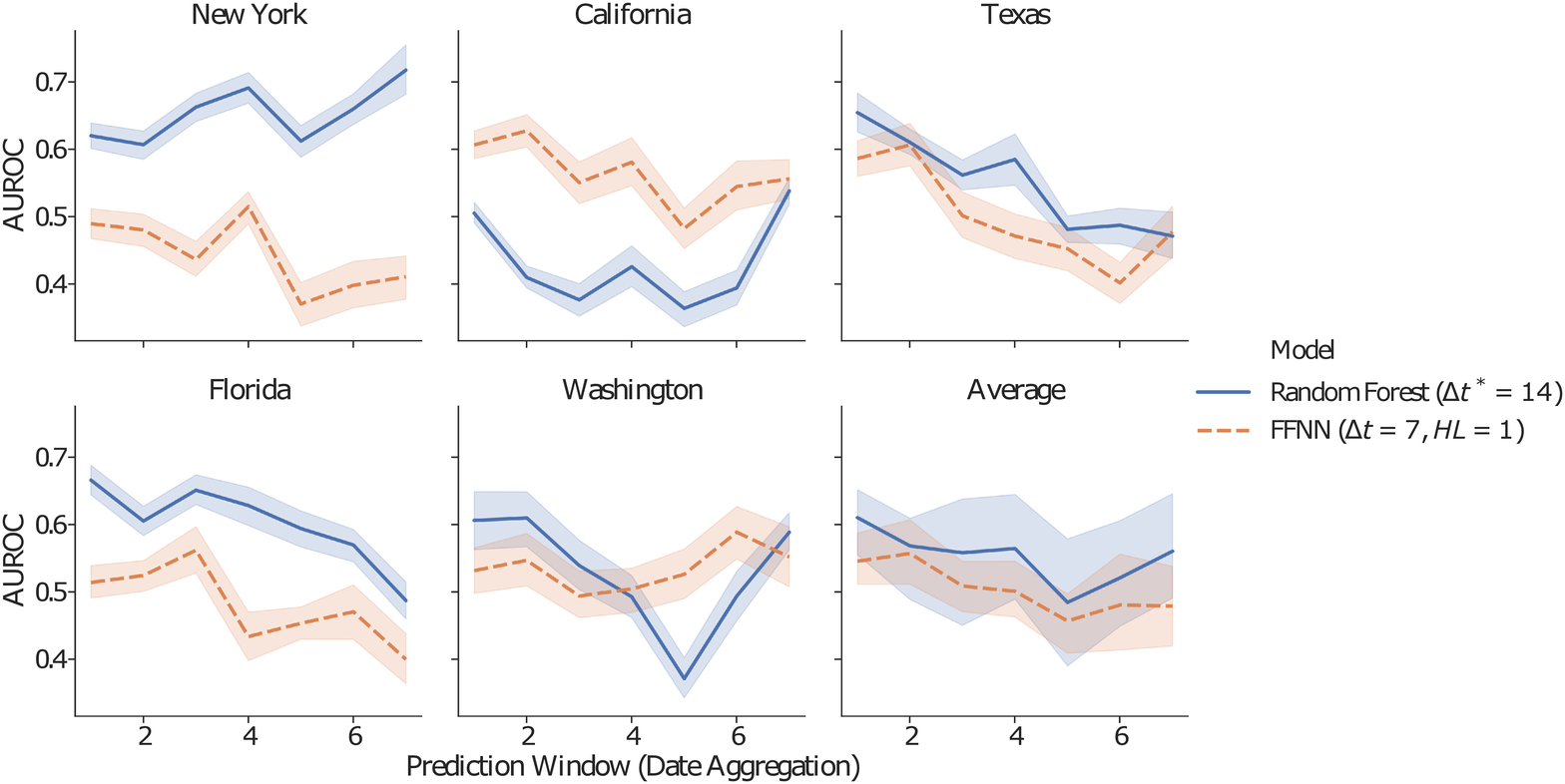}
    \caption{The reuslts of both label propagation (top) and date aggregation (bottom) show that using coarser date representations hinders the models' ability to discriminate between classes. In each plot, AUROC is shown as a function of $\Delta p$ (the prediction window) for the given state. Shaded regions represent the standard deviation across testing folds.}
    \label{fig:predwindows}
\end{figure}

Our baseline task is to predict the occurrence of a terrorist attack in a given state on a single day. Some prior work found success in making predictions over coarser periods of time, such as a week \cite{galla2018predicting}. However, we found this was not the case for our problem. We evaluated the following three methods for performing predictions over a given prediction window $\Delta p > 1$:
\begin{enumerate}
    \item Label propagation: we propagated event labels backward to the previous $\Delta p$ observations. For example, given an attack on Jan. 3 and $\Delta p=3$, we labeled Jan. 1, 2, and 3 as attacks.
    \item Date aggregation: we aggregated the news features from $\Delta p$ dates into a single observation. For example, given $\Delta p=3$, we combined Jan. 1, 2, and 3 into a single observation: Jan. 1-3. We further used mean pooling to generate the feature values for each aggregated observation.
\end{enumerate}

As shown in Figures \ref{fig:predwindows}, these approaches resulted in worse performance, with only a few exceptions. AUROC measures the trade-off between true positives and false positives, so whatever improvements larger prediction windows might have provided in successfully predicting attacks is counterbalanced by an increase in false positives. This is not surprising given that all of these methods run the risk of blurring the observed distinctions between classes. Date aggregation has the additional disadvantage of further reducing the number of training examples. A potential item of future work is to develop a more sophisticated method for making predictions over coarse periods of time without blurring the distinction between individual examples.

\subsection{Learning from Multiple States} \label{sec:transfer}

\begin{figure}
    \centering
    \includegraphics[width=0.366\textwidth]{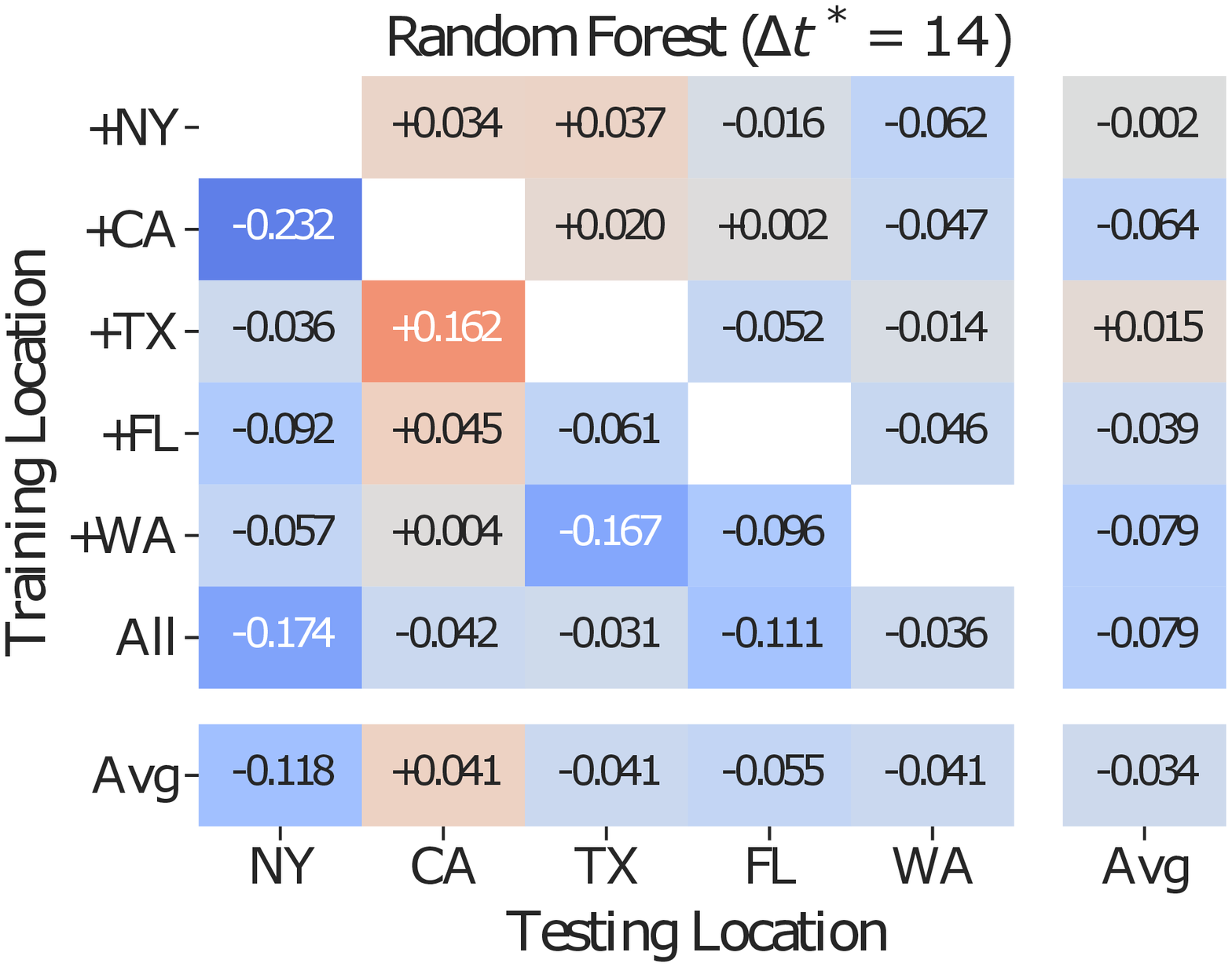}
    \hspace{0.05\textwidth}
    \includegraphics[width=0.436\textwidth]{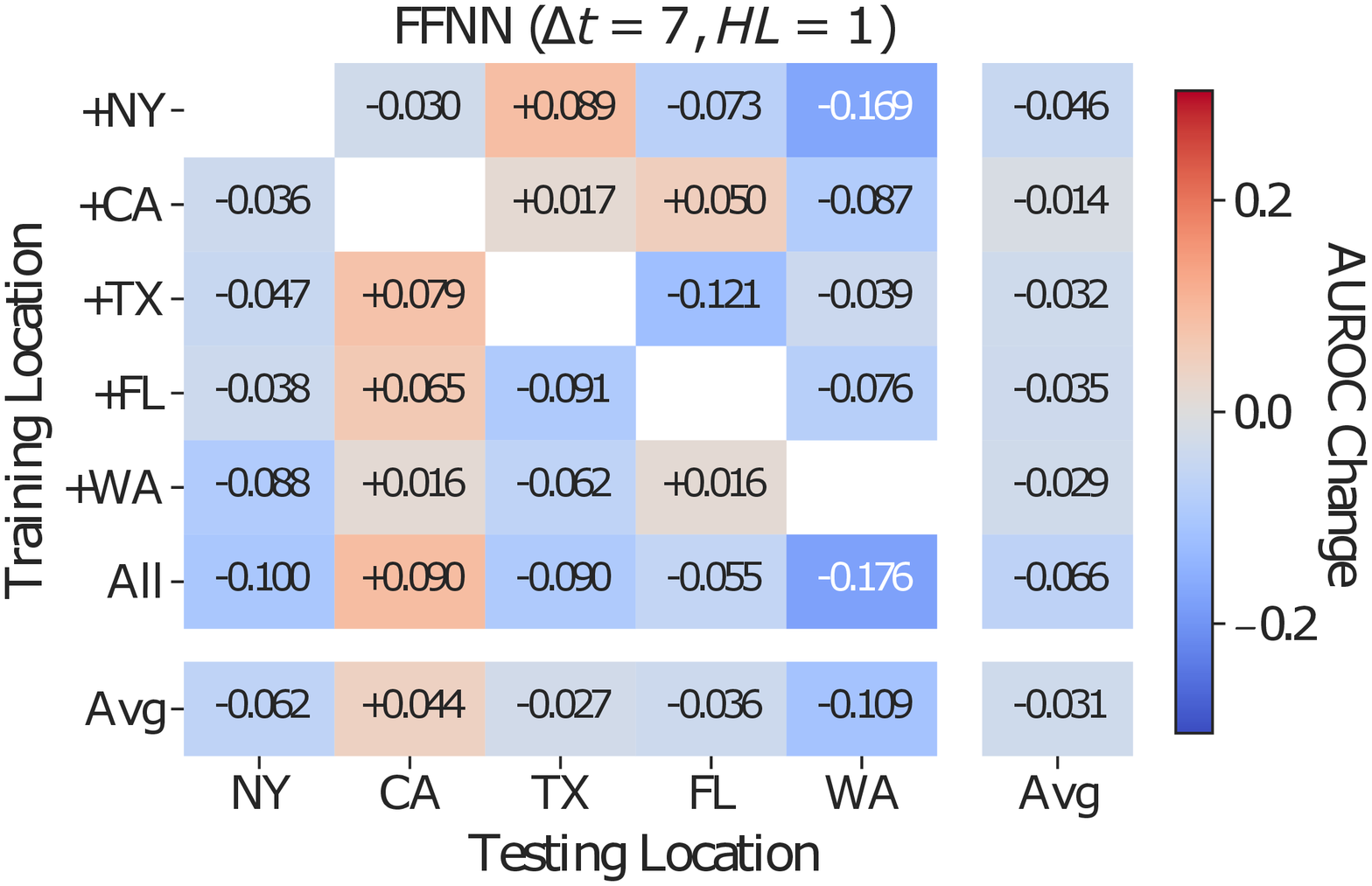}
    \caption{Training a model on data from multiple states reduced testing performance, on average, for both RF and FFNN. Each cell in the heatmap represents a unique model, with states on the x-axis representing the testing location and states on the y-axis representing the state that supplemented the training data. The value in each cell is the change in AUROC compared to the baseline model. For example, the cell (NY, +CA) represents a model tested on New York data and trained on a combination of New York and California data, and the value $-0.232$ means that this model produced an AUROC $0.232$ less than the baseline model that was only trained and tested on New York data (Table \ref{tab:resultsmain}). Reds and blues represent increases and decreases, respectively, to AUROC. Avg represents the average change in AUROC for the corresponding row or column.}
    \label{fig:tlocs}
\end{figure}

One of our key challenges is a lack of data. While our results thus far have shown that each state should be modeled independently, we also hypothesized that it could be possible to transfer knowledge between models. To evaluate the feasibility of such an approach, we trained a new set of models by combining data from multiple states. For example, when testing the New York model at baseline, we only included the training examples from New York; however, in these experiments we also supplemented the training set with additional data from another state (e.g., California) before evaluating on the New York testing set. As the heatmap in Figure \ref{fig:tlocs} shows, this only proved beneficial on average for California. This is not surprising given our previous results, which have suggested that the importance of feature groups and the optimal observation windows vary between states, and further evidences the need for models to be tailored to individual states.

\subsection{Predictions on Additional States}

\begin{table*}[]
    \centering
    \begin{tabular}{c|c|c|c}
         \multirow{2}{*}{\textbf{Testing State}} & \multirow{2}{*}{\textbf{Most Similar State(s)}} & \multicolumn{2}{c}{\textbf{Group Testing AUROC (5-fold cross-validation)}} \\
          & & Random Forest ($\Delta t^{*}=14$) & FFNN ($\Delta t=7$, $L=1$) \\ \hline
         LA & MO & $.417 \pm .149$ & $.605 \pm .076$ \\
         MO & KS, LA & $.466 \pm .175$ & $.601 \pm .097$ \\
         NV & UT, KY, MN & $.459 \pm .084$ & $.527 \pm .031$ \\
         PA & OH, IL, VA & $.500 \pm .164$ & $.491 \pm .164$ \\
         IN & OH, TN & $.520 \pm .110$ & $.575 \pm .090$ \\
         NC & VA, MD, NJ & $.548 \pm .280$ & $.654 \pm .214$ \\
         TN & KY, IN & $.461 \pm .108$ & $.419 \pm .113$ \\
         VA & NC, MD, NJ & $.548 \pm .280$ & $.417 \pm .151$ \\ \hline
    \end{tabular}
    \caption{Results for group testing on states that have experienced 5-7 attacks. The most similar state(s) are determined by hierarchical clustering on the news features and listed in order of similarity to the testing state.}
    \label{tab:statesimilarity}
\end{table*}

\begin{figure}
    \centering
    \includegraphics[width=0.8\textwidth]{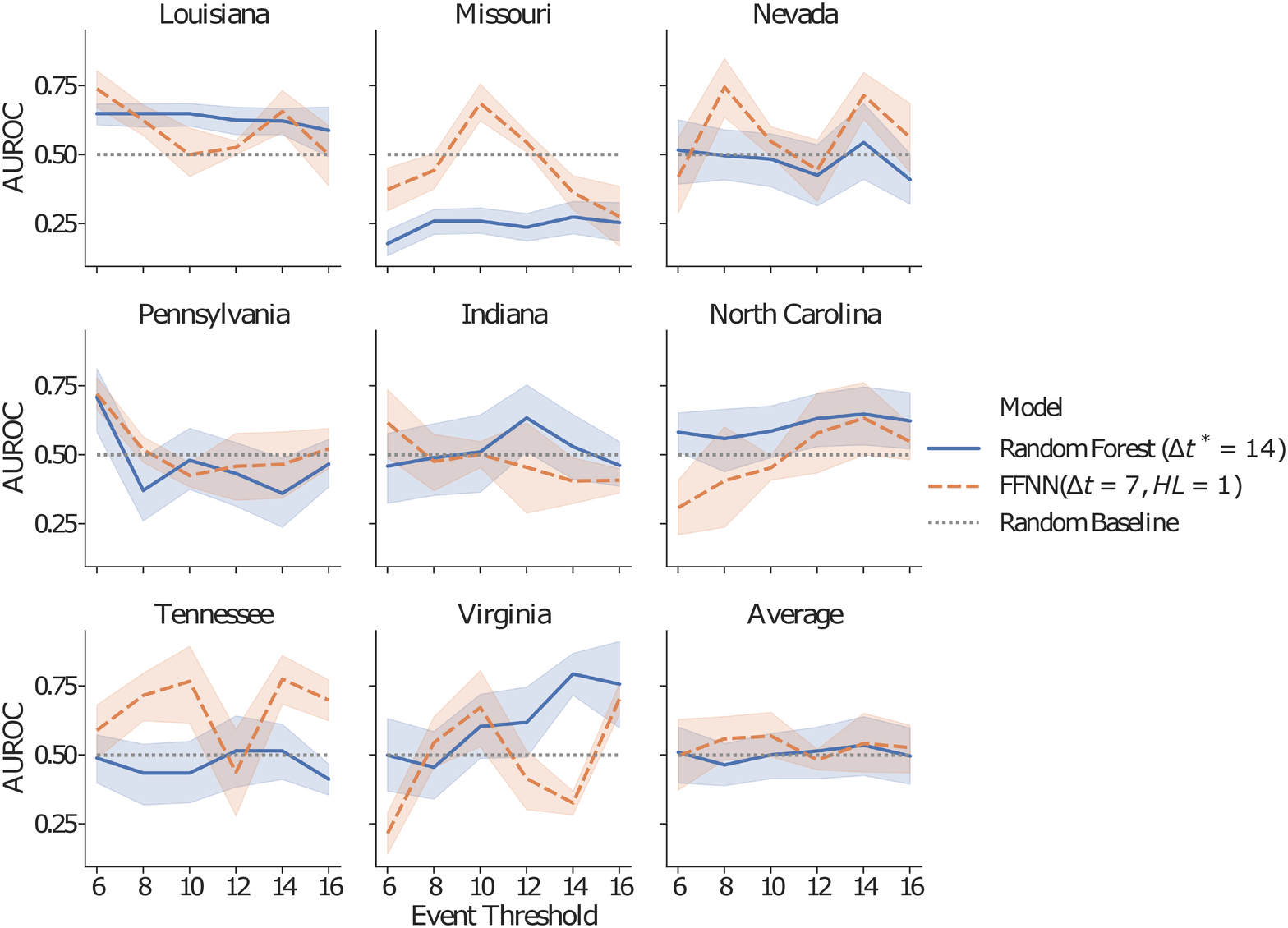}
    \caption{The results for single-state testing show that neither RF or FFNN perform significantly better than random on states that have experienced 5-7 attacks. Each plot shows AUROC as a function of the event threshold for each 5-fold cross validation experiment. The shaded areas represent the standard deviations of testing folds.}
    \label{fig:smlocs}
\end{figure}

So far we have limited our experiments and analysis to the five states with the most attacks. While we also wanted to evaluate our model on other states, none of them experienced more than seven recorded attacks (Table \ref{tab:attacksbystate}). We found that with such a small number of minority class examples it was impossible to learn an effective model. In an attempt to address this challenge, we considered the eight states that had experienced between five and seven attacks, and used hierarchical clustering (Euclidean distance, average link) to group them based on the observed news features. We tested two different grouping methods:
\begin{enumerate}
    \item Single-state testing. After choosing a single state to evaluate and splitting the data into training and test dates, we used hierarchical clustering to supplement the training set with data from the states most similar to the testing state. We continued to add additional states to the training set until the number of attacks in the training set exceeded a fixed threshold. In order to maintain proper training and testing separation, we only supplemented the training set with records that matched the dates of the records in the original training set. Then the model was trained on the multi-state training set and evaluated on the testing set from only the testing state (as in Section \ref{sec:transfer}). 
    \item Group testing. In these experiments, we also added data from the clustered states to the testing set, so that the states were evaluated as a group. For example, since Missouri (MO) is the state most similar to Louisiana (LA), we trained and tested the model on data from both MO and LA.
\end{enumerate}

Figure \ref{fig:smlocs} presents the results from single-state testing using event thresholds between 6 and 16. Table \ref{tab:statesimilarity} shows for each testing state the states that were most similar and thus clustered, along with the results of group testing. The inconsistent and overall poor performance of both single-state testing and group testing is not surprising given our findings in Section \ref{sec:transfer} that supplementing the training data with examples from additional states reduced model performance, even though in this case we only included the most similar states. These results suggest that in order to make meaningful predictions on these states, we either need more examples of attacks or a model that can identify and account for deeper differences between states.

%% file: conclusion.tex
\section{Conclusion} \label{sec:conclusion}

Terrorism presents a serious threat to human livelihood, even in the United States. In this work, we used a series of machine learning models trained on localized news data to predict the occurrence of terrorist attacks in a given state and on a given day---a problem that, to the best of our knowledge, no prior work has attemped to solve. Our best model---a Random Forest that uses a novel Kolmogorov-Smirnov moving average for feature representation---outperforms deep models and achieves an AUROC $\geq 0.667$ on four of five major states while being relatively insensitive to the amount of time between attacks. These results demonstrate that terrorist attacks can (to some degree) be predicted, and show the potency of a multimodal approach in using news data to characterize a spatiotemporal location. Our results further indicate that the key factors limiting model performance are noise and the small number of attacks in the data set. Finally, our results affirm the need for localized models; in our case, manifested by creating separate models for each state. We envision a number of areas for future work, including:
\begin{enumerate}
    \item Machine learning models that can better account for noise, either via manual feature engineering or noise-aware learning algorithms, and the differences between states.
    \item The application of news data to other multimodal learning problems.
    \item Continued efforts toward improving both the quality and quantity of terrorism-related data.
    \item Deeper studies into the relationship between news and specific terrorist attacks, such as investigating whether attacks can be traced to news topics like political events or socioeconomic themes.
\end{enumerate}
Continued research in the use of machine learning to predict terrorist attacks promises to further our understanding of terrorism, inform strategies for mitigating or even preventing the attacks, and save lives.